# A colour-marker tracker for pose estimation of slender surgical instruments: vision–language-model-guided adaptive colour calibration

Jaesung Hong[1,2,*] Dong Won Lee[3]

[1] R&D Center, A4LAB Inc., Gumi, Republic of Korea
[2] Department of Robotics and Mechatronics Engineering, DGIST, Daegu, Republic of Korea
[3] Department of Otorhinolaryngology–Head and Neck Surgery, Daegu Catholic University, Daegu, Republic of Korea

* Corresponding author: jhong@dgist.ac.kr, jhong@a4lab.co.kr

## Abstract

**Background.** Slender surgical instruments need small, narrow markers for camera-based pose estimation. A colour-stripe marker fits this requirement, but its colours in the image drift from the design values with lighting, printing and camera settings. Because the colours can only be corrected after the marker has been found, methods that rely on the design colours face a circular problem. This study proposes vision–language-model-based adaptive colour calibration and compares it with rule-based methods.

**Methods.** Three methods were compared. Two rule-based methods re-measure the colours from detections made with the design colours: Initial colour update does so once after tracking starts, and Colour adaptation continuously during tracking. VLM guidance uses a fine-tuned vision–language model (VLM) to find the markers without relying on the design colours and builds the profile from their pixels. Once the colour profile is obtained, all three share the tracker, the final detection condition and the evaluation criteria. Recall, F1 score and false positives were compared on 672 synthetic image pairs, eight degraded laboratory videos and a surgical-robot (da Vinci) video.

**Results.** Recall/F1 (%) for Initial colour update, Colour adaptation and VLM guidance were 92.0/95.7, 87.8/93.3 and 98.2/99.1 on the synthetic images; 51.3/67.2, 48.8/65.0 and 83.8/87.0 on the laboratory videos; and 38.9/55.2, 19.7/29.7 and 100/100 on the surgical-robot video, where the colours were strongly shifted. In that video the rule-based methods never found the purple/cyan marker, whereas VLM guidance found all 234 markers with no false positives. In a simulation of a two-stripe marker on a shaft, the hue boundary between the stripes moved by less than 0.1 pixel for exposure gains of 0.5–2.0, compared with 2.9–4.0 pixels for a boundary found with a fixed brightness threshold, and it matched the stability of the best black-and-white boundary method, the gradient extremum.

**Conclusions.** When the degradation is mild, the rule-based methods, which need no training and no GPU, are sufficient. When the colours change greatly, the gain comes from a VLM-guided candidate search that does not rely on the design colours, that is, from breaking the circle. Combining colour-stripe markers with adaptive colour calibration enables tracking of slender surgical instruments with small markers.

## 1 Introduction

Many instruments whose position and orientation need to be tracked in the operating room are slender: endoscopes, trocars and cannulas, forceps and electrocautery tools. Their shaft diameter is around 5–15 mm, and a marker attached at the handle end is far from the instrument tip, so a small pose error is amplified at the tip like a lever. In transoral robotic thyroidectomy (TORT), trocars of about 10 mm diameter are inserted through three incisions inside the lower lip, a little over 3 cm apart, and their motion is transmitted directly to the lip and to the mental nerve [1,2]. Tracking such instruments with a camera requires a marker that is small, narrow and easy to attach along the shaft (Figure 1).

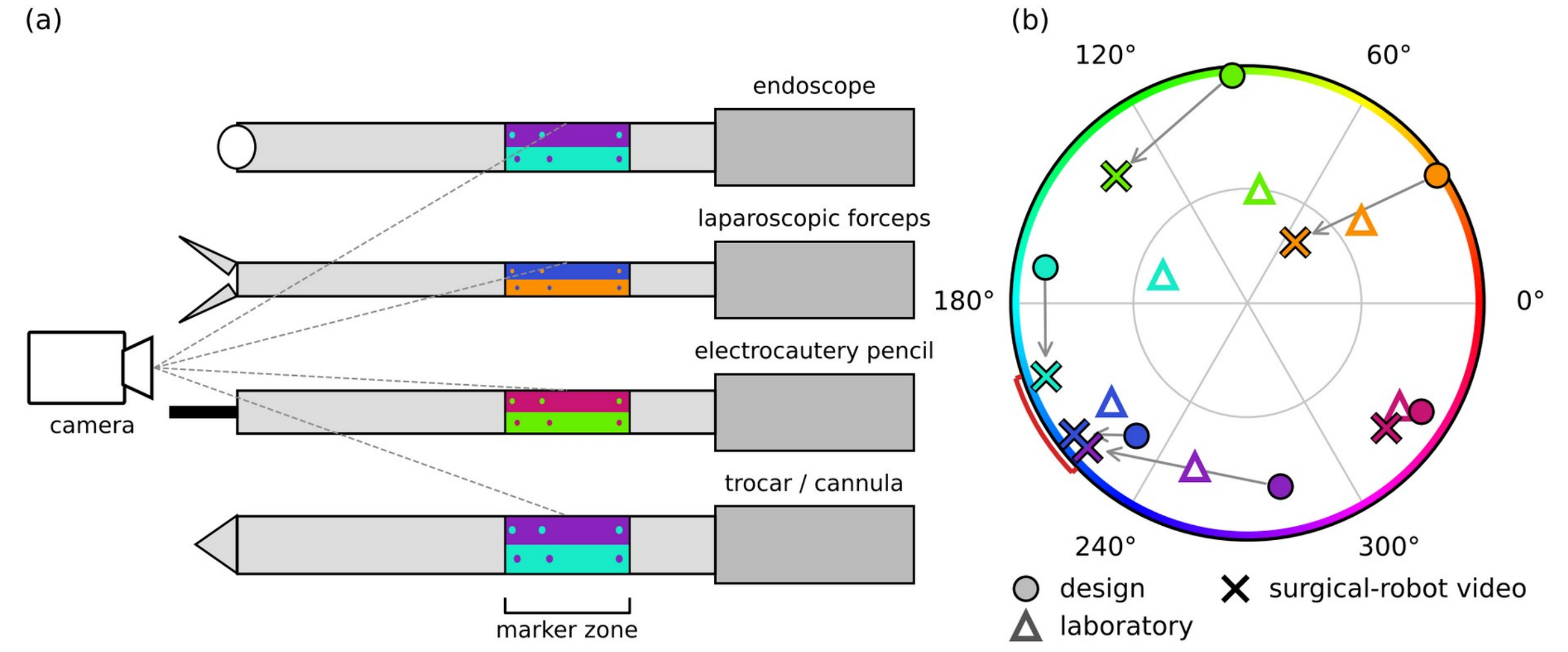

**Figure 1.** Marker positions on slender surgical instruments and the colour shift of the marker inks. (a) Positions on slender surgical instruments where a marker can be attached, and the external camera. (b) Design hue–saturation of the six inks used in this study (circles) and the values measured in laboratory images (triangles) and in the surgical-robot video (×). The laboratory values of the blue/orange pair were measured in a photograph from the previous study [15]. The red bracket marks the two inks of the purple/cyan marker (CS) in the surgical-robot video.

The most widely used fiducial markers are black-and-white square patterns such as ArUco [3] and AprilTag [4,5], which are robust to brightness changes [5,6]. On a narrow, round shaft, however, the width of a square pattern and its cell size are limited and the curved projection must be taken into account [7], although black-and-white markers for cylinders have been proposed [8–10].

Distinguishable colours can add identification information to a spatial pattern [11,12], and colour has been used to segment laparoscopic instruments [13] and to make markers that contrast with tissue [14]. The dual colour stripe (DCS) marker that we invented [15] uses the boundary between two colour stripes as a long centre line and distinguishes orientation and ID with dots inside the stripes.

An unavoidable problem with colour markers is that colour is not a fixed value: it changes with the ink, the illumination and the camera settings [16,17]. When a marker with the purple/cyan pair was used in a video that we recorded during surgical-robot operation, the two stripe hues, about 110° apart in the design, were only about 20° apart in the video (Figure 1(b)), and a tracker with fixed design colours never found this marker. This problem is circular. To correct colours that have drifted from the design, the marker must first be found and its pixels read, but a detector with fixed design colours cannot find a marker whose colours have drifted.

This paper proposes adaptive colour calibration, which automatically updates the colour profile to the environment, in a rule-based form and in a form based on a vision–language model (VLM). The contributions are threefold. First, we examine the cues provided by the marker structure and by the colour information in a slender mounting area, and compare by simulation the boundary stability of black-and-white and colour two-stripe markers. For pose precision, we analyse the structural differences caused by the arrangement and number of features separately from colour itself. Second, we implement two rule-based methods, which are referred to as Initial colour update and Colour adaptation. They update the colour profile from detections made with the design colours according to rules defined by a person. We also propose an alternative VLM-based method (hereafter VLM guidance), which uses a trained VLM to find the markers without relying on the design colours and thereby breaks this circle. Third, we compare the three methods as complete automatic pipelines on synthetic, laboratory and surgical-robot videos.

## 1.1 Related work

Fiducial markers include black-and-white square patterns [3–5,18,19], topological and circular patterns [20] and colour patterns [11,12,21], and learning has been used to overcome difficult lighting [22] (Table 1). Surgical instruments are tracked with marker-less [23] or marker-based approaches [8,10,14,24–26]. Among the latter, printed patterns viewed with an ordinary camera are smaller and more affordable than dedicated optical trackers that use retro-reflective markers [27], but they are sensitive to lighting. Near-infrared retro-reflective markers [25] and square markers combined with an RGB-D camera [26] are accurate but need separate illumination or sensors and a flat place to attach a planar marker. Few marker studies focus on the colour change of colour markers itself. Cho and Neumann [21] proposed detection rules invariant to brightness, and Wei et al. [13] proposed learning the colours before surgery. To our knowledge, no study has yet used a vision–language model to find markers and calibrate their colours.

**Table 1.** Comparison of related work. The colour-change handling column describes how each method responds when the colour seen by the camera differs from the design value.

| Study | Marker form | Use of colour | Curved surface | Colour-change handling | Application |
|---|---|---|---|---|---|
| ArUco [3], AprilTag [4,5], ARTag [18] | black-and-white square grid | none | planar assumption | not applicable (adaptive threshold) | general |
| TopoTag [20] | black-and-white topological pattern | none | planar assumption | not applicable | general |
| ChromaTag [11] | square grid of opponent colours | faster detection | planar assumption | fixed colours assumed | general |
| Colour barcode [12], multi-ring fiducial [21] | colour cells, colour rings | more information | planar assumption | brightness-invariant rules | general, AR |
| DeepArUco++ [22] | black-and-white square grid | none | planar assumption | lighting handled by learning | general |
| Zhang et al. [8] | black-and-white dots and chessboard vertices | none | cylinder | not applicable | ultrasound probe |
| CylinderTag [9] | black-and-white cyclic pattern | none | cylinder | not applicable | general |
| Gadwe and Ren [10] | green stripe and black-and-white square modules | segmentation by green stripe | cylinder | fixed colours | endoscopic instruments |
| Cartucho et al. [14], Huang et al. [24] | dot pattern on green background | separation from background | cylinder | fixed HSV range | laparoscopic instruments, gamma probe |
| Wei et al. [13] | colour markings | instrument segmentation | not applicable | colour learning before surgery | laparoscopic visual servoing |
| DCS [15], this study | two colour stripes and dots | ID and centre line | cylinder (curvature used) | profile update by real-time pixel sampling | surgical instruments |

# 2 Materials and methods

## 2.1 Colour-stripe marker

The dual colour stripe (DCS) marker is a thin rectangular tag (Figure 2(a)) [15]. Its most important design features are that the long side lies along the shaft axis and that two nearly complementary colour stripes meet at the centre line. The colour pair can be chosen for the application; this study used three pairs, purple/cyan (CS), blue/orange (RS) and magenta/green (TS). Each stripe carries three dots in the colour of the opposite stripe at asymmetric spacing, which tells the order of the dots and the orientation of the marker. For a cannula of 10 mm diameter, the whole marker is reduced to 32 × 9.6 mm.

The width and height of the marker can be scaled up or down depending on the object to which it is attached, and their ratio can also be changed. On a da Vinci cannula of 10 mm diameter, a single 32 × 9.6 mm film marker gives a six-degree-of-freedom pose. The design is based on the assumption that in many operations the tilt of the instrument axis and the position of the instrument tip matter more than the rotation about the instrument axis. Three dots per stripe were chosen so that the pose can still be determined by PnP from four dots when two of the six are possibly lost. Even when some dots disappear, the colour pair identifies the marker and the dot spacing identifies the stripe. On a shaft of 15 mm diameter, the 12 mm high marker wraps about 90°. Joining four such markers gives a wrap-around marker that encircles the shaft.

The centre line is the longest feature and stably constrains the projected axis direction, while the dots and the outer edges constrain the remaining components, including roll (Section 3.1). The contribution of curvature in marker design was treated in another paper [15]; this paper concentrates on colour, especially real-time calibration of colour. The roles of the stripe structure and of the colour information were compared with black-and-white markers in simulations of the boundary position as the exposure changes and of the pose precision in the same mounting area (Sections 2.2, 2.8 and 3.1).

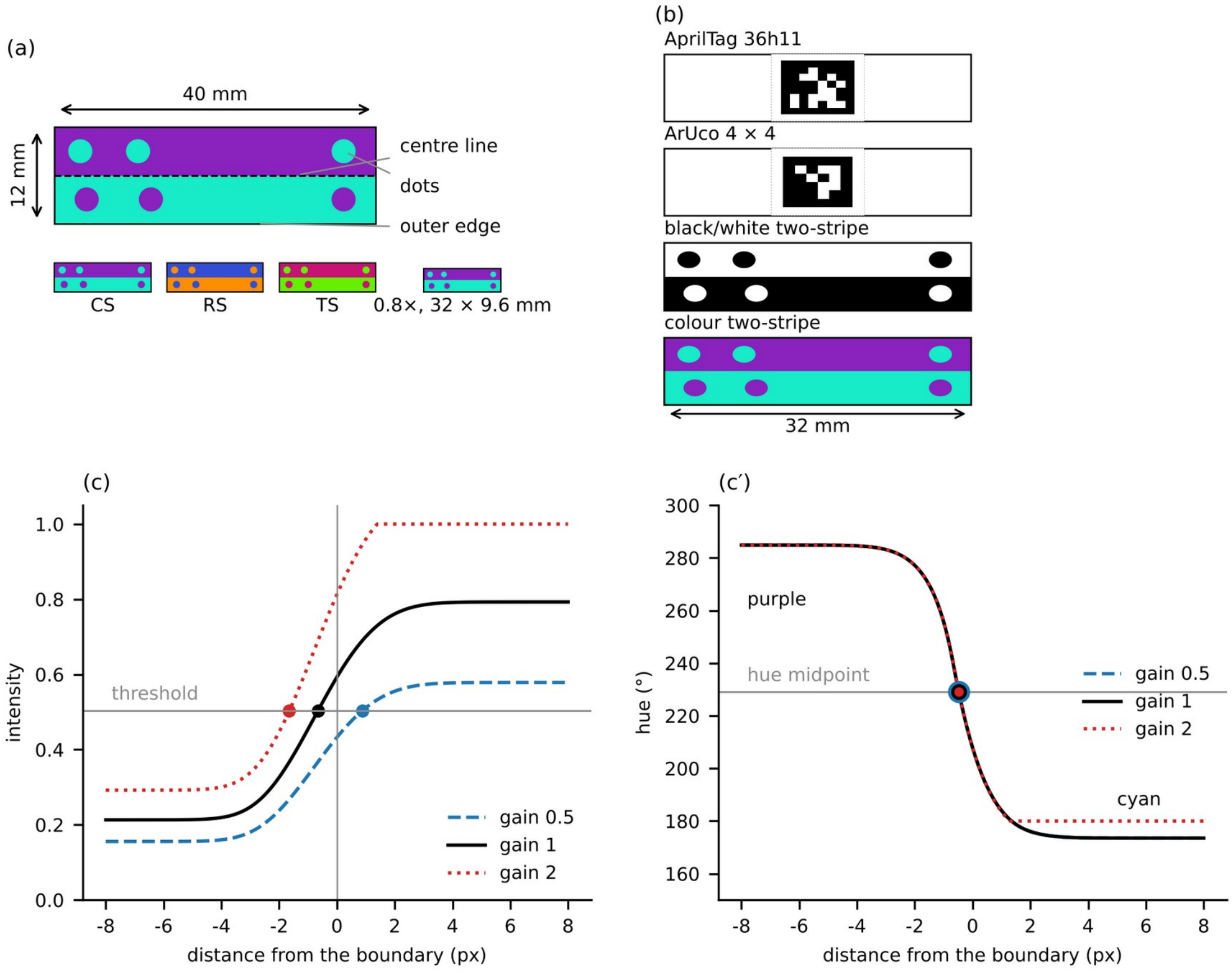

**Figure 2.** Colour-stripe marker. (a) A single 40 × 12 mm marker (centre line, dots, outer edges) with the three colour pairs (CS, purple/cyan; RS, blue/orange; TS, magenta/green), and the 0.8× version for a cannula of 10 mm diameter. (b) Four markers that fit the same 32 × 9.6 mm area. (c) Brightness profile of a black-and-white boundary and (c′) hue profile of a colour boundary at camera gains of 0.5, 1 and 2. The dots mark the crossings of a fixed threshold in (c), which move with the gain, and of the hue midpoint in (c′), which coincide; the quantitative comparison with the black-and-white adaptive threshold and gradient methods is given in Section 3.1.

## 2.2 Marker structure and colour information on slender instruments

Black-and-white markers have the advantages of large luminance contrast, no need for colour calibration and mature well-known detectors. However, on a slender instrument, the mounting area, the minimum feature size, the boundary extraction method and the separability from the background must be considered together. We distinguish the roles of the stripe structure and of the colour information from the following three viewpoints.

First, the minimum feature size and the ID capacity within the mounting area. If an $n \times n$ cell tag is placed in an area of width $w$, one cell side including the quiet zone is $c = w/(n + 2)$, and the cell size seen by a camera of focal length $f$ at distance $Z$ is as follows.

$$p = \frac{f\,c}{Z} \tag{1}$$

With $w$ = 9.6 mm and AprilTag 36h11 ($n$ = 8), $c$ = 0.96 mm, and with $f$ = 1458 pixels and $Z$ = 300 mm, $p$ = 4.7 pixels. Once blur and noise are added, reading the cells becomes difficult. A colour stripe in the same area appears 23 pixels wide when the stripe width of 4.8 mm is substituted for the cell side in Equation (1) (1458 × 4.8 / 300), and a large part of the ID is carried by the colour pair of the stripes rather than by cells (Figure 2(b)). The large features of stripes and dots also apply to a black-and-white two-stripe marker, and colour uses the distinguishable colour pair as an additional identification cue.

Second, the position of the boundary. The boundary between the two stripes is the centre line that strongly constrains the pose, so a shift of the boundary becomes a pose error. When a black-and-white boundary is found with a fixed brightness threshold $T$, and the brightness near the blurred boundary is written as $I(x) = g\,s(x)\,\rho(x)$ ($g$ is the exposure gain, $s$ the shading and $\rho$ the blurred reflectance), the crossing point satisfies $g\,s(x)\,\rho(x) = T$ and therefore moves with $g$ and $s$. A black-and-white boundary can, however, also be found with a local adaptive threshold or with a gradient extremum, and these methods reduce the shift caused by exposure changes. A colour boundary can use the hue difference independently of brightness, and hue is expressed as a relative value between channels as follows.

$$H = \operatorname{atan2}\left(\sqrt{3}\,(G - B),\, 2R - G - B\right) \tag{2}$$

In a model in which the same factor $g\,s(x)$ multiplies all three channels, the hue of Equation (2) does not change [16,17]. Therefore, as long as sufficient saturation is maintained and the effects of saturation clipping and noise are small, hue provides a boundary cue that is less sensitive to brightness changes. This property is admittedly limited when the illumination colour, the white balance, the per-channel non-linear processing or the clipping changes. We found that the ink colours already differ from the design values on several kinds of printer, and that under operating-room lighting the hue difference between the two stripes shrinks greatly. The design values of the six inks used in this study and the values measured in laboratory images and in the surgical-robot video are given in Figure 1(b) and Table 2. Figure 2(c) and (c′) show an example that compares a fixed brightness threshold with hue at camera gains of 0.5, 1 and 2. A quantitative comparison that also includes the black-and-white adaptive threshold and gradient methods is given in Table 3. The proposed colour calibration is needed to reflect the hue and the saturation and value distributions of the input data.

**Table 2.** Measured values of Figure 1(b). Each cell gives hue (degrees) / saturation (0–255); the design profile is the value in the marker definition file, and the other values are the medians of the inner stripe pixels in the corresponding image. The photograph is that of the 15 mm shaft from the previous study [15], the laboratory video is the pre-degradation frame of Figure 4 (640 × 360 reduced copy), and the surgical-robot video is frame 105 in Figure 4 (original resolution). The values measured in frame 495 agreed with frame 105 within 2–4° of hue. As laboratory values, Figure 1(b) shows those of the laboratory video for the purple/cyan and magenta/green pairs and those of the photograph for the blue/orange pair.

| Ink (marker) | Design profile | Photograph of the previous study | Laboratory video | Surgical-robot video |
|---|---|---|---|---|
| purple (CS) | 280° / 208 | – | 252° / 192 | 222° / 241 |
| cyan (CS) | 170° / 230 | – | 162° / 100 | 200° / 240 |
| blue (RS) | 230° / 193 | 216° / 188 | – | 217° / 243 |
| orange (RS) | 34° / 255 | 36° / 157 | – | 52° / 86 |
| magenta (TS) | 328° / 229 | 344° / 146 | 326° / 205 | 318° / 208 |
| green (TS) | 94° / 255 | 82° / 158 | 84° / 128 | 136° / 204 |

Third, separability from the background. Surgical images contain shadows, specular highlights, gauze, metal shafts and tissue together [28]. Two adjacent stripes of different colours can provide an additional cue that is difficult to obtain from a single colour or from luminance alone [11]. The rarity of a particular colour pair and its advantage in detection depend on the scene. This study, however, did not directly compare the number of candidates or the false-detection rate of black-and-white and colour detectors over whole surgical backgrounds, so the background separability of a colour pair is not interpreted as a general advantage over black-and-white markers.

## 2.3 Colour profile

The tracking software classifies pixels into one of the two stripes in HSV space. The profile of stripe $i$ consists of the medians and tolerances of hue, saturation and value, $\{H_i, S_i, V_i, \Delta H_i, \Delta S_i, \Delta V_i\}$. The stripe colour $c_{\text{obs}}$ seen by the camera can be written as follows.

$$c_{\text{obs}} = \Gamma\left(M \cdot \text{diag}(\ell) \cdot \rho_{\text{print}}\right) \tag{3}$$

$\rho_{\text{print}}$ is the reflectance of the printed ink, $\ell$ the per-channel intensity of the illumination, $M$ the colour transform and white balance of the camera, and $\Gamma$ the tone curve and compression. All four factors change in the field, so the profile must be re-estimated from the current video. The task of colour calibration therefore reduces to one thing: finding the pixels that actually belong to each stripe without knowing the colours.

When the hues of the two stripes are far enough apart, pixels are classified with a hue window ($|H - H_i| \leq \Delta H_i$, $S \geq S_{\min}$); when they are close, pixels are assigned to the stripe with the smaller HSV distance.

## 2.4 Initial colour update (rule-based)

In this paper, rule-based means that the first candidate is found by the design-colour detection rules of the marker definition rather than by a learned model. Initial colour update, the first of the two rule-based methods, attempts detection with the design profile within the first 10 s after tracking starts, with the design colour windows widened slightly. From every detection whose pose the tracker judges usable for measurement, the inner stripe pixels excluding the dots and edges are collected. The marker centre of each detection is checked for continuity with the last approved frame before its pixels are collected. When the collection ends, a new profile is built from the medians and spreads of these pixels for each marker with at least 40 pixels per stripe and installed in the tracker immediately; a marker with insufficient samples keeps the design profile, and its collection is retried automatically. The method rests on two assumptions: that the camera is usually set up so that the markers are close and clearly visible when tracking starts, and that under such conditions the initial detection succeeds even if the recorded colours differ slightly from the design colours. It is simple because it ends with a single calibration at the start, but it may fail when the set-up at the start is poor.

## 2.5 Colour adaptation (rule-based)

Colour adaptation, the second rule-based method, updates the profile in small steps whenever a detection that passes the quality conditions occurs during tracking. The updated profile is validated with a shadow run, and a watchdog reverts it if detections decrease. The method is intended to follow slowly changing illumination or a moving auto white balance. The two rule-based methods share the same limitation: a first detection with the design profile is needed before any pixels are available for colour calibration. When the colours deviate only slightly, the widened colour windows allow a first detection, but when they deviate greatly, as for the CS marker in the surgical-robot video, neither method obtains a starting point.

### 2.6 VLM guidance (learning-based)

VLM guidance, the main proposal of this study, finds the markers without relying on the design colours, in order to break the contradictory circle in which the design values are consulted to find colours that have departed from those values (Figure 3). The model receives one image and a short instruction and returns the rectangle and the ID of each marker candidate. The instruction for locating the markers describes only the appearance, ‘a flat or instrument-wrapped rectangular printed marker with two parallel colour stripes and small contrasting circular dots’, and contains no colour names, which prevents the model from taking a shortcut through the design colours. The marker definition is used only for the stripe correspondence, the meaning of the IDs and the profile rules, not for finding the first candidate. The absence of colour names in the instruction does not mean that the model uses no colour information from the image. The model looks at colour and structure together; the difference is that it does not filter the first candidates with a fixed HSV range as the rule-based methods do. The description of the appearance in the instruction serves to draw on the language prior knowledge of the model.

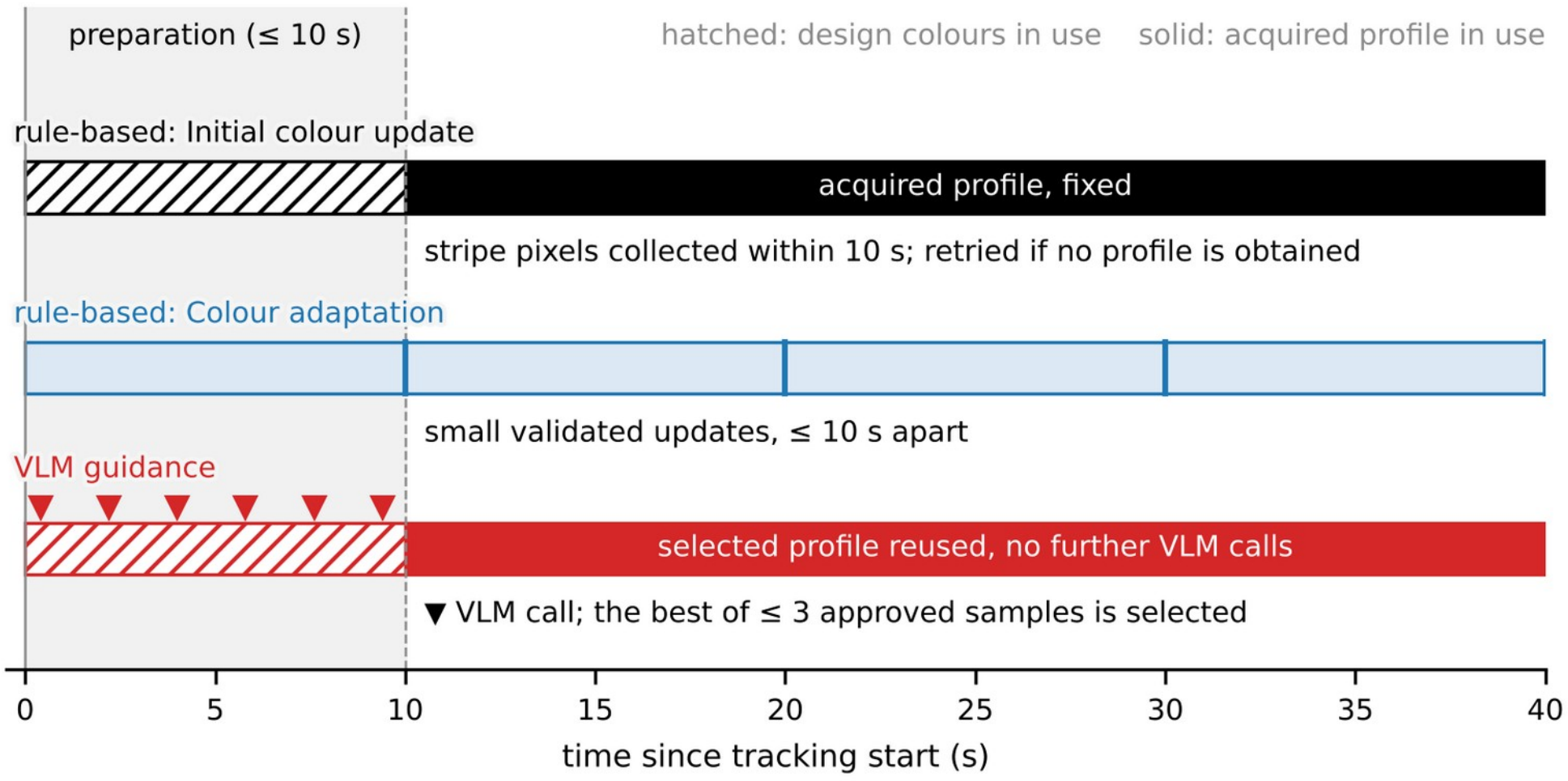

**Figure 3.** When the three methods obtain and apply the colour profile. Hatched sections use the design colours, and filled sections use the profile acquired from the video. VLM, vision–language model.

*Model.* The base model is Qwen3-VL-4B-Instruct [29], quantised to 4 bits [30], and it was trained and run on a single GPU with 8 GB of memory (NVIDIA GeForce RTX 4070 Laptop GPU). This open model is small enough to run on a computer at the site, so surgical video does not have to leave the hospital, which matters for the security of medical data. LoRA [31] was used for fine-tuning because it can be trained within 8 GB, adds almost no inference cost and is the most widely validated method of its kind. It was attached only to the language-model layers. A location adapter proposes the rectangle of each candidate, and an ID adapter re-judges the ID from the relative appearance of the two stripes. The ID adapter does not require all inner dots to be visible. The input was limited to about 0.6 megapixels, and greedy decoding ensures the same answer for the same frame.

*Training data.* The adapters were trained without manual labelling, on laboratory images labelled automatically by the existing tracker and on their degraded versions [32–34]. In clear laboratory recordings, the positions, outlines and IDs of markers that the existing tracker found reliably were used as automatic labels (pseudo-labels), and these labels were passed unchanged to degraded versions of the same frames to which colour neutralisation, exposure and channel gain, blur, resolution reduction, noise and JPEG compression were applied in stages. This is because, even when the existing detector misses a marker in a degraded image, the position and ID obtained from the clear original remain valid. The training data of the location adapter are 2,424 rows, the 606 originals plus three degraded sets of 606 each; this number is not the number of independent recordings but the number of training rows including crops and degraded derivatives. The degradation ranges were derived from the colour design values in the marker definition files and did not use values from the evaluation videos. The development data are 365 degraded rows from separate laboratory recordings with non-overlapping capture groups. None of the evaluation data, including the 672 synthetic evaluation pairs and the surgical-robot video, was used for training.

*Colour sampling and profile.* When exactly one candidate of an ID exists in a frame, geometry code that uses no colour prior finds the outline and the centre line of the two stripes and reads the original pixels in four 3 × 3 pixel sample windows on each stripe. The sample is approved when the window colours agree and both stripes are supported along the centre line, without pose estimation or a condition on the number of inner dots. Up to three approved samples per marker are collected within 10 s of video time after the first approval. Each is tested by installing its profile in the tracker and detecting the marker in its own frame (native detection support), and the best-supported sample is selected and fixed. The profile is built with the same rule as for Initial colour update, except that a sample with too low a saturation is applied by a hue-only classifier, and it is reused in later frames without further VLM calls.

The four sample windows on each stripe give the colours of 36 original pixels per stripe. Which of the two stripes is the upper stripe of the design is decided by comparing the two correspondence hypotheses of the marker definition, and the decision is deferred when ambiguous. More precisely, the sample is approved when the colours of the sample windows agree with each other, both stripes are supported over at least three quarters of the centre-line section, and the fraction of empty sections is 0.15 or less. These are the conditions of the sample approval stage, and they are distinguished from the final marker detection condition applied after the calibrated profile is installed. The choice among the approved samples is made after the third approval or 10 s of video time after the first approval, whichever comes first; when the detection support is equal, the sample of higher quality and then the earlier one is chosen. Colours from several times are not averaged, and no sample is chosen by looking at later recognition rates. When the saturation of the selected sample is too low to meet the installation condition, the profile is applied by a circular hue classifier inside the same tracker, and the pixel eligibility of this path is looser than that of the rule-based methods. The final detection applies the tracker and the geometric validation conditions common to the three methods.

*Choice of a VLM.* A VLM was chosen instead of a CNN detector for a reason confirmed in a preliminary experiment. During development, a Faster R-CNN [35] trained on the same synthetic training data found 46 of 60 markers in 20 real frames and produced 37 spurious rectangles. The VLM adapter found 57 and produced no spurious rectangles. This development comparison only shows that the chosen VLM configuration gave better results on real frames than the compared Faster R-CNN configuration; the contributions of language prior knowledge [36], visual pre-training, model size and training settings were not separated.

## 2.7 Data and evaluation

The synthetic images are 672 pairs produced by applying 21 degradation conditions to 32 virtual print lots: an undegraded reference, 17 single degradations (printing deviation, warm or cool lighting, exposure, channel mixing, noise, blur, JPEG compression) and three combinations. Each pair consists of an image for acquiring the colours and an image for reusing the acquired profile, with three markers each (2,016 markers). The laboratory videos are eight videos made by applying four conditions of optical degradation and colour neutralisation to two recordings of markers on a shaft of 15 mm diameter (80 markers). The surgical-robot video (Figure 4) was recorded in an operating room with an external HD camera, with a da Vinci system docked to a TORT training mannequin; it shows 32 × 9.6 mm markers (CS, RS, TS) on three trocars of 10 mm diameter (234 markers). This video was not used for model or parameter tuning. All input images have a resolution of 1920 × 1080.

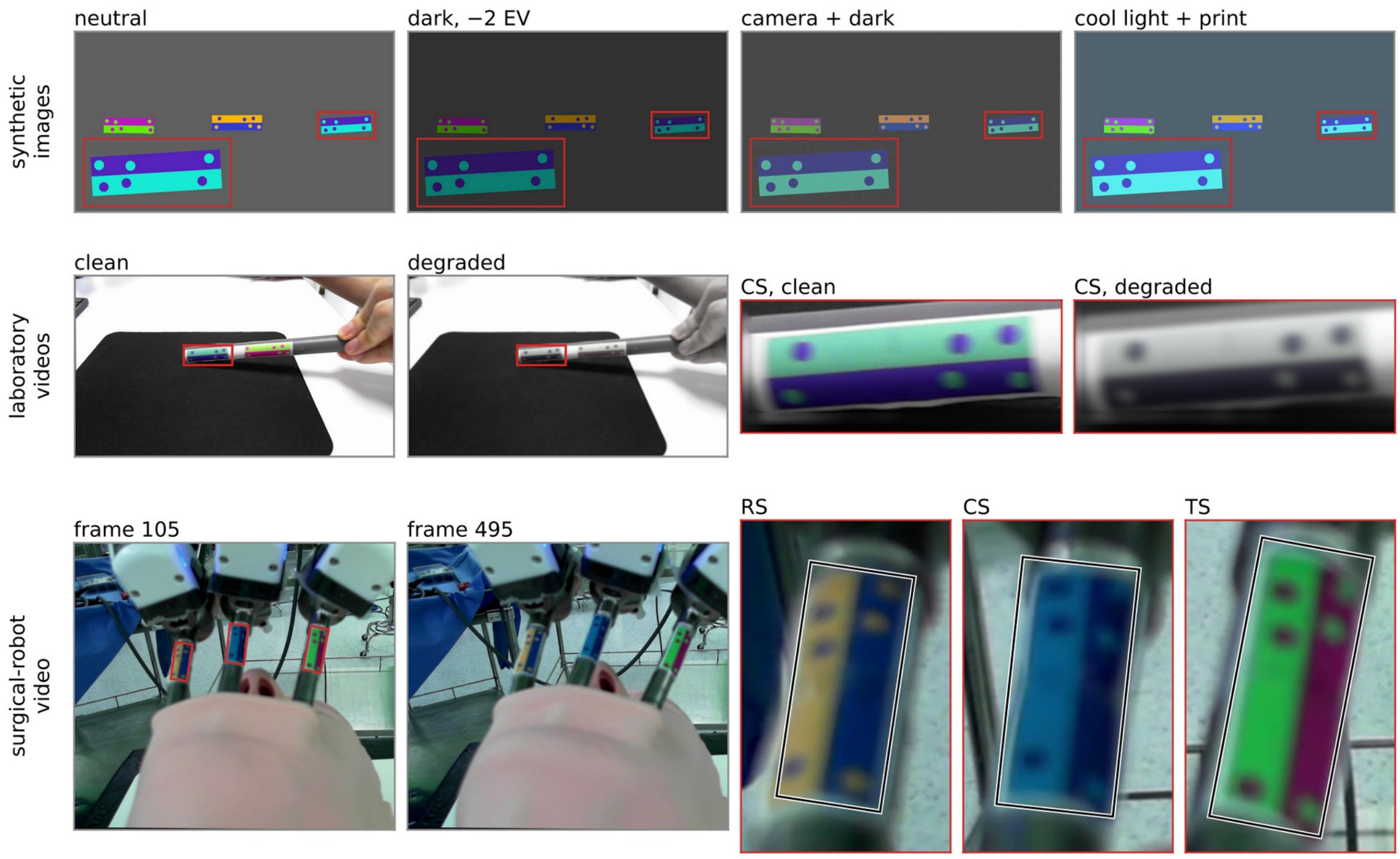

**Figure 4.** Examples of the evaluation images. None of the data sets was used for model training. Top: four conditions of the synthetic images with the CS marker enlarged. Middle: original and degraded frames of a laboratory recording with the CS marker enlarged. Bottom: two frames of the surgical-robot video and enlargements of the three markers in frame 105. The lines on the surgical-robot images mark the marker outlines. CS, purple/cyan marker; RS, blue/orange marker; TS, magenta/green marker; EV, exposure value.

The 21 conditions include seven strong single degradations, and their parameters are listed with the results by condition in Section 3.3. Acquisition and reuse must be evaluated on different images to tell whether the profile fits only one image. In Figure 4, the synthetic images are the reference condition, strong under-exposure, a combined camera degradation and a combined cool-lighting and printing condition; the laboratory images are the original and a strongly colour-neutralised and optically degraded version of the same frame; and the surgical-robot images are frames 105 and 495. The surgical-robot video consists of 880 frames extracted at 5 Hz, and the 78 frames with references were evaluated.

A detection with the same ID as a reference marker and a centre within 20–40 pixels of it, depending on the marker size, was counted as a true positive (TP); unmatched references are false negatives (FN) and the remaining detections false positives (FP). The recognition rate is reported as recall, TP/(TP + FN), together with the F1 score, 2TP/(2TP + FP + FN), which also reflects false positives.

The centre tolerance is $r = 0.30\sqrt{A}$ limited to 20–40 pixels, where $A$ is the area of the reference marker on a 1920 × 1080 scale, so it does not depend on the size of the detected box. Each reference is matched to at most one detection of the same ID. The references of the laboratory and surgical-robot videos were annotated with AI assistance and partly checked by the authors against the original videos. This check is a verification step for the evaluation results and is distinguished from the execution procedure that acquires marker positions and colour samples automatically.

The three methods were compared as complete automatic colour calibration pipelines that start without human operation, each with a preparation period of up to 10 s. What was unified is the processing after the colours are obtained: the tracker code, the marker definitions and camera calibration, the final detection condition, the profile-building rule, and the evaluation frames and criteria. The same model weights, instructions and settings were used for all data sets. The colour calibration stage itself differs by the definition of each method. Processing time was measured for all methods on the same computer. The same six input videos were given to each of the three methods, and the runs were repeated three times.

After the accuracy evaluation was completed, the runs were executed sequentially on the same computer (Intel Core Ultra 9 185H, NVIDIA GeForce RTX 4070 Laptop GPU) with OpenCV fixed to one thread, and only the time of the tracker API calls was measured, excluding the time for saving the evaluation records to files. For VLM guidance, the time of the first colour acquisition, with the model already loaded on the GPU, is reported separately from the per-frame processing after profile reuse.

### 2.8 Simulations of colour and black-and-white markers

The comparison of colour and black-and-white markers was made with two simulations (Tables 3 and 4). In the boundary simulation, the position bias of black-and-white boundaries (fixed and adaptive thresholds, gradient maximum) and of the colour boundary (hue) was measured on a two-stripe marker on a shaft of 15 mm diameter while the exposure gain and the illumination direction were varied. In the pose simulation, the pose precision of a square tag (four corners) and of a two-stripe marker (dots, outline corners, centre line, outer edges) in the same available mounting area was compared by Monte Carlo simulation. The Gaussian noise per feature was $\sigma$ 0.2 pixels for the tag corners and $\sigma$ 0.3 pixels for the dots, outline corners and line samples of the two-stripe marker. This is a structural comparison that includes the marker size and the feature arrangement, and it does not isolate the effect of colour itself. The detailed conditions are as follows.

*Boundary simulation (Table 3, Figure 5).* A 40 × 12 mm two-stripe marker (without dots) was placed on a cylinder with a print-surface radius of 7.64 mm and rendered by ray tracing, in which the intersection of the ray of each pixel with the cylinder was computed analytically. The true boundary was obtained analytically by projecting the surface line at roll $\psi$. The camera is $f$ = 1167 pixels at a distance of 150 mm and $f$ = 1458 pixels at 100 mm, roll 0° and 20°, blur $\sigma$ 1.0 and 1.5 pixels, Bayer interpolation, noise $\sigma$ 3 and 20 seeds. The illumination is a point source at the camera position (L1) and a parallel light 45° above the optical axis (L2), and the exposure gain $g$ has eight levels from 0.25 to 2.8 (a frontal white surface is 200/255 at $g$ = 1). The reflectance of the black-and-white stripes is 0.90/0.05, and the colour stripes use the linearised sRGB values (237, 28, 36)/(0, 162, 232) of the red/cyan pair (a representative pair, not the purple/cyan pair used in the evaluation of this paper). Each boundary rule was applied column by column, a straight line was fitted over the whole marker length, and the offset at the centre was measured. The rules are a fixed threshold of 128, a local (max + min)/2, a vertical-derivative extremum (parabolic interpolation), the hue mask gap midpoint (design hue ±20°, $S \geq 60/255$, $V \geq 30/255$, gap ≤ 6 pixels) and the hue zero crossing (sub-pixel zero crossing of the difference of the angular distances to the two design hues, $S \geq 25/255$). The roll conversion is the small-angle approximation $\delta\psi \approx (\delta x \cdot Z / f) / R$. Because the colour rules use the design hues as they are, profile errors and white balance changes are not included.

*Pose simulation (Table 4, Figure 6).* Two markers were placed in the same 32 × 9.6 mm area on a cylinder of radius 5 mm, their features were projected from the true pose (marker centre on the optical axis, distances 100–400 mm, roll 0° and 20°), Gaussian noise was added, and the pose was re-solved by non-linear least squares (Levenberg–Marquardt, initial value near the truth). The square tag was modelled with the four corners of the black border of a 9.6 mm tag (a 7.68 mm square) at their true coordinates on the cylinder, and the planar-model variant solved the same curved observations with planar coordinates at $z$ = 0. This approximation differs from a control experiment with a tag actually attached to a plane, and the differences in dimensions and projection model between the curved and planar coordinates may be included in the error. The two-stripe marker is 32 × 9.6 mm with six dots (upper stripe −13.4/−7.7/+12.8 mm, lower stripe −12.8/−6.4/+12.8 mm, ±2.4 mm in the arc direction), four outline corners (±16 mm, ±4.8 mm), 40 points on the centre line and 20 points on each outer edge (within ±15.2 mm). For points on lines, only the residual normal to the projected line was used. The noise $\sigma$ is 0.2 pixels for the tag corners, 0.3 pixels for the dots and outline corners and 0.3 pixels for the points on lines, with 300 repetitions per condition. The roll error is the component of the error rotation about the cylinder axis, and the axis-direction error is the angle between the estimated and true axes.

# 3 Results

## 3.1 Colour versus black-and-white simulations

Table 3 shows how the boundary position of a two-stripe marker on a shaft of 15 mm diameter changes with the exposure gain $g$. A black-and-white boundary found with a fixed threshold moved by 2.9–4.0 pixels for $g = 0.5$–2.0 depending on the illumination direction, and for $g \leq 0.35$ the white stripe fell below the threshold and no boundary was found. In this geometry one pixel corresponds to a roll of about 0.96°, so this shift corresponds to about 2.8–3.8°. Over the same gain range, the variation of the local adaptive threshold was 0.25–0.28 pixels and that of the gradient extremum was 0.01–0.09 pixels. The hue mask midpoint varied by 0.08 pixels and the hue zero crossing by 0.01 pixels, so the gradient method was about as stable as the hue methods. These results therefore show that hue is stable compared with a fixed threshold, but they do not mean that it is superior to all black-and-white boundary extraction methods. The changes under extreme exposure and saturation conditions are shown in Figure 5. The constant offset of a boundary and its variation with gain must be distinguished. This simulation is for the representative red/cyan pair and the specified illumination model, and it does not include white balance changes or real colour calibration errors.

**Table 3.** Position bias of the boundary between the two stripes (pixels) and the corresponding roll error (°). Two-stripe marker of 40 × 12 mm on a shaft of 15 mm diameter at a distance of 150 mm. Illumination L1 is a point source at the camera position and L2 is a parallel light 45° above the optical axis. Positive values are biases towards the dark stripe, and one pixel corresponds to a roll of about 0.96°. The variation is the difference between the maximum and the minimum for exposure gains $g = 0.5$–2.0.

| **Boundary rule** | $g = 0.5$ | $g = 1.0$ | $g = 2.0$ | **Variation, L1** | **Variation, L2** | $g \leq 0.35$ |
|---|---|---|---|---|---|---|
| Black-and-white, fixed threshold | −1.10 (−1.1°) | +0.68 (+0.7°) | +1.81 (+1.7°) | 2.91 (2.8°) | 3.95 (3.8°) | fails |
| Black-and-white, adaptive threshold | +0.78 (+0.8°) | +0.78 (+0.8°) | +1.03 (+1.0°) | 0.25 (0.2°) | 0.28 (0.3°) | works |
| Black-and-white, gradient maximum | +0.79 (+0.8°) | +0.79 (+0.8°) | +0.88 (+0.8°) | 0.09 (0.1°) | 0.01 (0.0°) | works |
| Colour, hue mask gap midpoint | +0.35 (+0.3°) | +0.40 (+0.4°) | +0.43 (+0.4°) | 0.08 (0.1°) | 0.08 (0.1°) | works |
| Colour, hue zero crossing | +0.27 (+0.3°) | +0.26 (+0.3°) | +0.26 (+0.3°) | 0.01 (0.0°) | 0.01 (0.0°) | works |

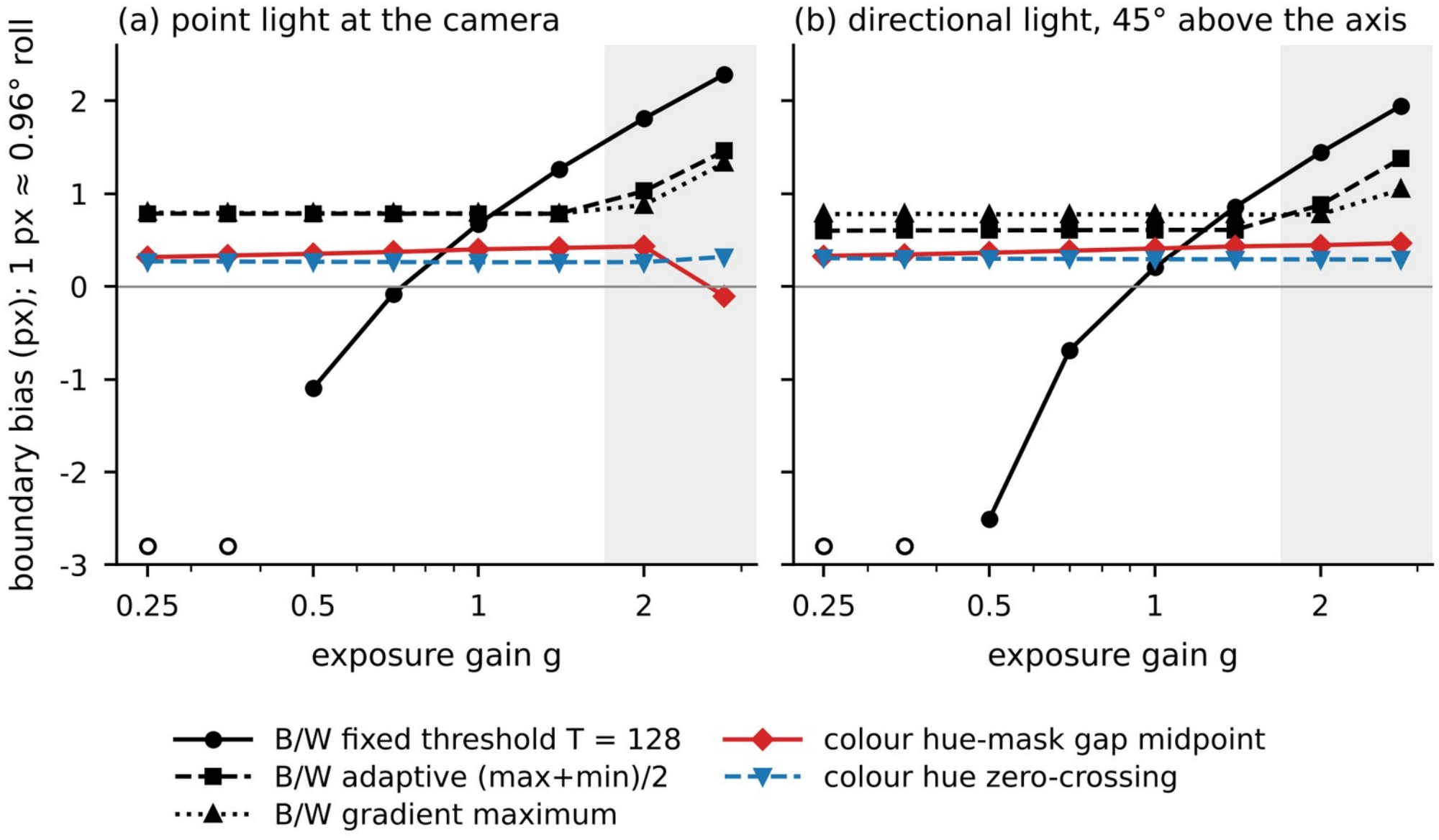

**Figure 5.** Position bias of the boundary between the two stripes as a function of exposure gain (raw data of Table 3). (a) Point source at the camera position, (b) parallel light 45° above the optical axis. Open circles mark cases without a threshold crossing, and the grey band marks $g > 1.71$, where the white stripe saturates.

Table 4 gives the pose precision for different feature configurations in the same available mounting area. Near a frontal view, the roll RMS of the four corners of a square tag modelled with the cylindrical geometry was 3.0–6.9° at distances of 10–40 cm, 0.4–1.9° when the six dots and four outline corners of the two-stripe marker were used, and 0.1–0.5° when the centre line and outer edges were added. This difference reflects together the long 32 mm structure, the depth distribution and number of the features and the specified position noise. It is not a result that isolates an improvement of pose precision by colour itself. At roll 20°, the corners of the square tag separate in depth and its roll RMS fell to 0.5–1.6°, while the two-stripe marker using the dots, centre line and outer edges gave 0.1–0.4°. These results are for the specified feature arrangement and the Gaussian position-noise model, and they were not validated against feature detection failures in real images or against correlated line-sample errors.

**Table 4.** Pose precision comparing the feature configurations of a square tag and of the two-stripe structure in the same 32 × 9.6 mm available mounting area. Each cell gives roll RMS (°) / axis-direction RMS (°) / position RMS (mm), frontal view (roll 0°), 300 trials. The feature position noise $\sigma$ is 0.2 pixels for the tag corners and 0.3 pixels for the dot, corner and line samples of the two-stripe marker. The planar-approximation row is the result of fitting curved observations with planar coordinates and does not mean that planar markers are generally unable to estimate the pose.

| **Features (pose model)** | **10 cm** | **20 cm** | **30 cm** | **40 cm** |
|---|---|---|---|---|
| Square tag, 4 corners (cylindrical model) | 3.0 / 2.8 / 0.23 | 4.7 / 4.8 / 0.92 | 6.0 / 5.8 / 2.3 | 6.9 / 6.9 / 3.9 |
| Square tag, 4 corners (planar approximation) | 25 / 0.3 / 1.6 | 25 / 0.7 / 1.6 | 25 / 1.0 / 2.2 | 25 / 1.2 / 3.2 |
| Two-stripe, 6 dots + 4 outline corners | 0.42 / 0.25 / 0.05 | 0.90 / 0.70 / 0.20 | 1.4 / 1.2 / 0.49 | 1.9 / 1.6 / 0.79 |
| Two-stripe, dots + centre line + outer edges | 0.11 / 0.22 / 0.04 | 0.25 / 0.61 / 0.17 | 0.35 / 1.0 / 0.40 | 0.48 / 1.6 / 0.69 |

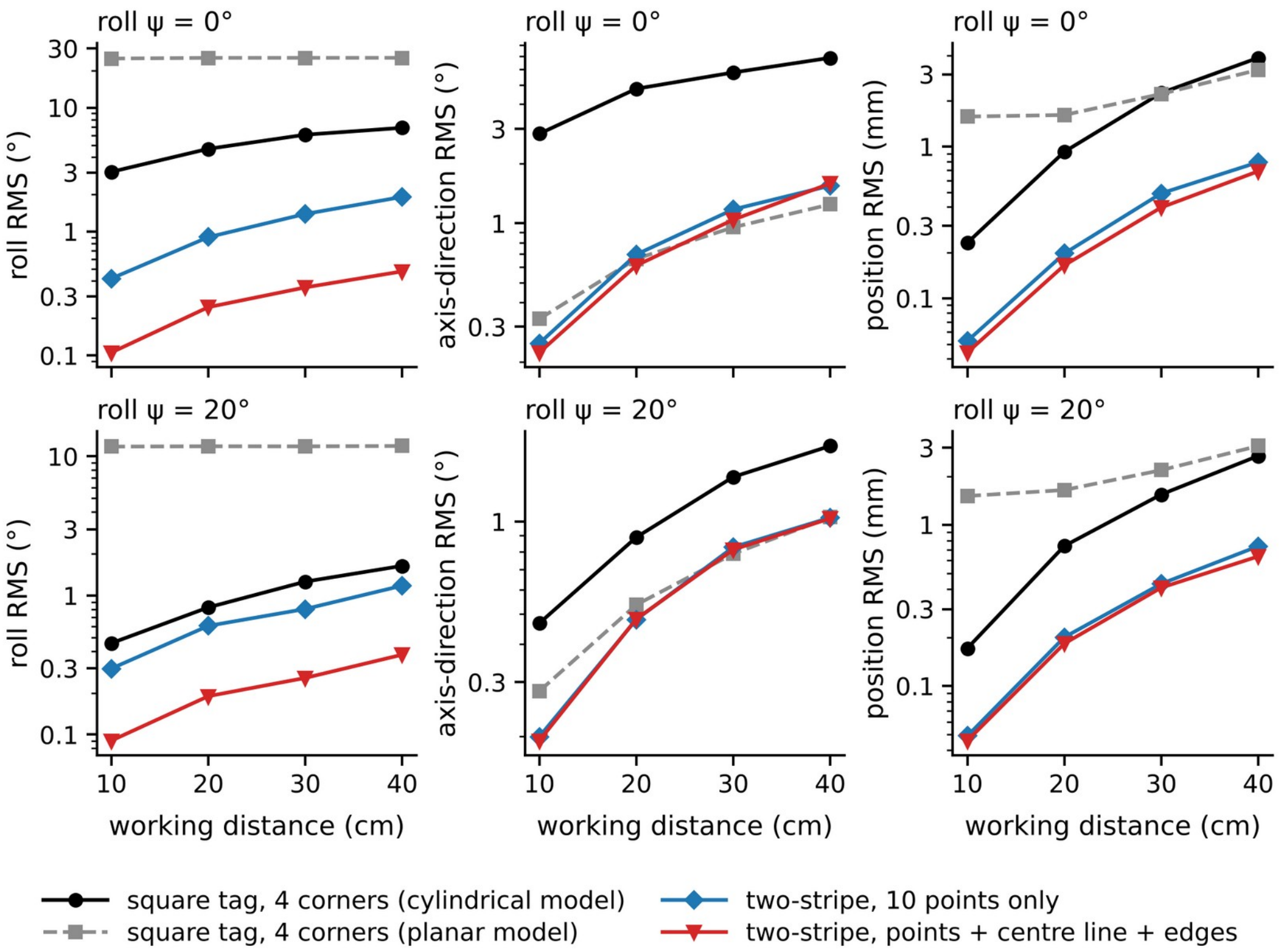

**Figure 6.** Pose precision of a square tag and of a two-stripe marker in the same 32 × 9.6 mm area (raw data of Table 4). Roll, axis-direction and position RMS by distance at roll 0° and 20°. The square tag and the two-stripe marker differ in the number and arrangement of features and in the position noise conditions. The grey planar-model curves are approximate results obtained by applying planar coordinates to curved observations and do not isolate the effect of colour or an intrinsic limitation of planar markers.

## 3.2 Recognition performance of the three methods

On the synthetic images, recall and F1 were 92.0% and 95.7% for Initial colour update, 87.8% and 93.3% for Colour adaptation, and 98.2% and 99.1% for VLM guidance (Table 5); the results by condition are given in Section 3.3. On the laboratory degraded videos the differences widened. Recall and F1 were 51.3% and 67.2% for Initial colour update and 48.8% and 65.0% for Colour adaptation, compared with 83.8% and 87.0% for VLM guidance. The 7 false positives of VLM guidance were cases in which a non-marker region was taken as a marker; each rule-based method had 1. Under strong colour neutralisation the two rule-based methods could not obtain a first detection and remained at the design colours.

**Table 5.** Recognition performance of the three colour calibration methods. Each cell gives recall / F1 (%), and N is the number of evaluated markers. The last column gives the numbers of CS, RS and TS markers found in the surgical-robot video (78 each). The numbers of false positives are given below the table.

| Method | Synthetic images, 672 pairs (N = 2,016) | Laboratory degraded videos, 8 conditions (N = 80) | Surgical-robot video (N = 234) | CS / RS / TS markers found in the surgical-robot video (78 each) |
|---|---|---|---|---|
| Rule-based: Initial colour update | 92.0 / 95.7 | 51.3 / 67.2 | 38.9 / 55.2 | 0 / 18 / 73 |
| Rule-based: Colour adaptation | 87.8 / 93.3 | 48.8 / 65.0 | 19.7 / 29.7 | 0 / 17 / 29 |
| VLM guidance | 98.2 / 99.1 | 83.8 / 87.0 | 100 / 100 | 78 / 78 / 78 |

A marker counts as found when a detection with the same ID lies within $r = 0.30\sqrt{A}$ of the reference centre ($r$ limited to 20–40 pixels; $A$ is the area of the reference marker on a 1920 × 1080 scale). Numbers of false positives (synthetic / laboratory / surgical-robot video): Initial colour update 4 / 1 / 5, Colour adaptation 7 / 1 / 30, VLM guidance 0 / 7 / 0. Of the 30 false positives of Colour adaptation on the surgical-robot video, 26 were detections wrongly labelled CS. CS, purple/cyan marker; RS, blue/orange marker; TS, magenta/green marker; VLM, vision–language model.

On the surgical-robot video, where the colours were strongly shifted, recall and F1 were 38.9% and 55.2% for Initial colour update and 19.7% and 29.7% for Colour adaptation, and neither method ever found the CS marker (Table 5). VLM guidance found all 234 markers with no false positives. Initial colour update found 0, 18 and 73 of the 78 CS, RS and TS markers, and Colour adaptation found 0, 17 and 29; of the 30 false positives of Colour adaptation, 26 were detections wrongly labelled CS. An earlier diagnosis showed that such false detections pointed at the blue surgical drape on the left of the image.

## 3.3 Synthetic images by condition

All three methods found all 96 markers in 15 of the 21 conditions, and the differences arose in the conditions that change the colours (Table 6, Figure 7). Under conditions that deviate slightly from the design colours, the first detection succeeds with the design colours and the subsequent update fills in the rest, so the rule-based methods are sufficient. Under strong warm lighting the two rule-based methods found 64 of 96 markers and VLM guidance 94. Under strong cool lighting, Initial colour update found 49 of 96 markers, Colour adaptation 18 and VLM guidance 94. Under strong channel mixing only Colour adaptation fell, to 65, a case of an update frozen in the wrong direction. Under moderate cool lighting (75 of 96) and under the combined condition of under-exposure, channel mixing, noise and JPEG (93, with 3 false positives), Colour adaptation was the least stable. The one condition that all three methods found difficult was the combined condition of cool lighting, printing deviation and channel mixing, in which the rule-based methods found 15 markers and VLM guidance 64. In this condition, it appears that the cool lighting moved the purple stripe towards blue and the printing deviation and channel mixing lowered the saturation of both stripes, so that the two stripe colours could not be separated even when the position was found (Figure 4). All false positives of the two rule-based methods (4 and 7) came from strong cool lighting and the two combined conditions.

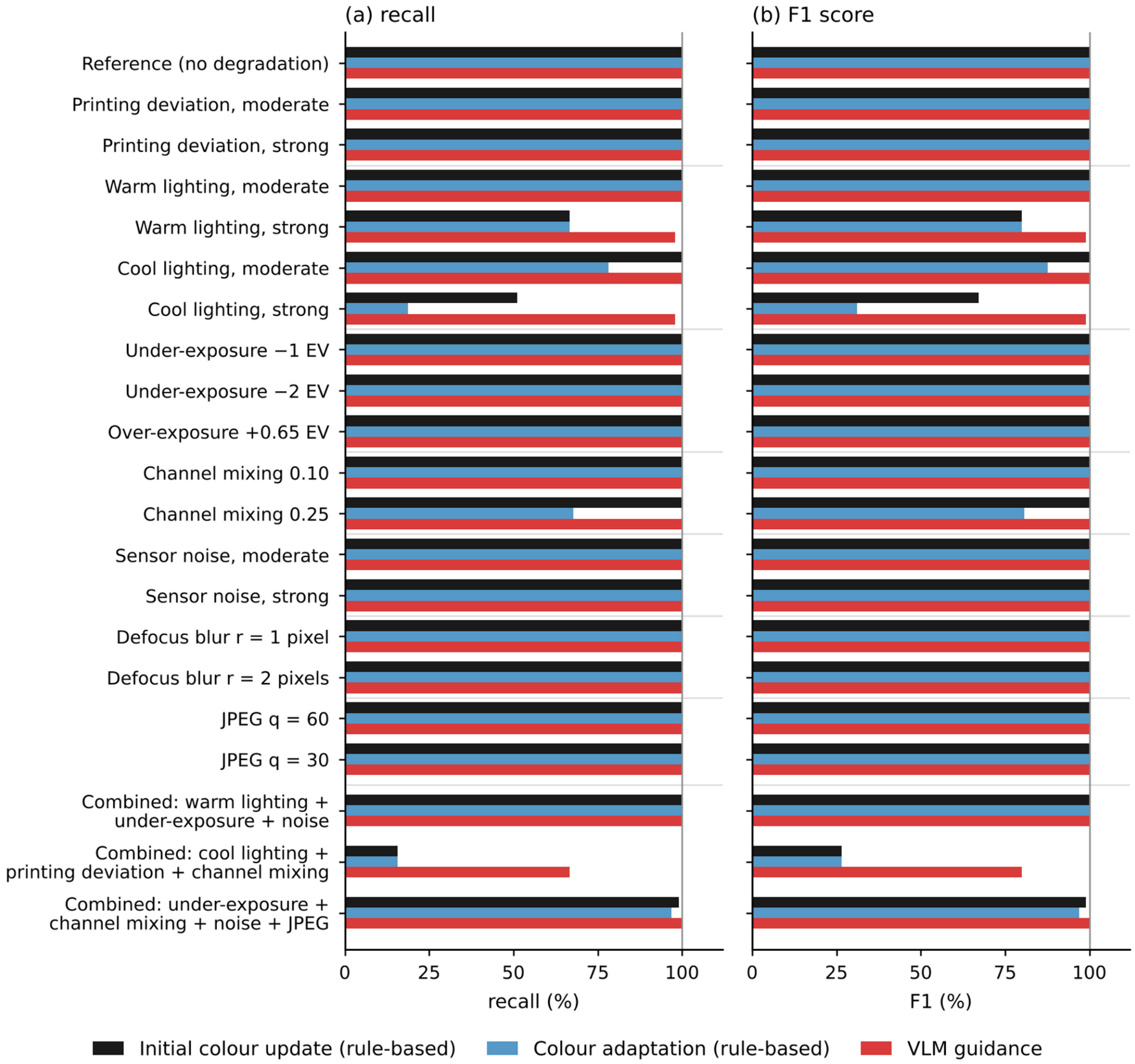

**Figure 7.** Results on the synthetic images by degradation condition. (a) Recall, (b) F1 score; each condition contains 96 markers. The numbers of markers found and of false positives are given in Table 6.

**Table 6.** Results on the synthetic images by degradation condition. Each cell gives the number of markers found (of 96) and recall / F1 (%), with the number of false positives in parentheses. The condition parameters are the illumination colour cast (RGB gains 1.18/1.00/0.82 and 1.38/1.00/0.62, reversed for the cool cast), printing deviation strengths 0.04, 0.16 and 0.32, channel mixing 0.10 and 0.25, noise (1,500 photoelectrons with 3 e read noise, and 400 with 7 e) and JPEG quality 60 and 30. Because the 32 print lots have a repeated structure, small differences between conditions are not interpreted statistically.

| Condition | Initial colour update (rule-based) | Colour adaptation (rule-based) | VLM guidance |
|---|---|---|---|
| Reference (no degradation) | 96; 100 / 100 | 96; 100 / 100 | 96; 100 / 100 |
| Printing deviation, moderate | 96; 100 / 100 | 96; 100 / 100 | 96; 100 / 100 |
| Printing deviation, strong | 96; 100 / 100 | 96; 100 / 100 | 96; 100 / 100 |
| Warm lighting, moderate | 96; 100 / 100 | 96; 100 / 100 | 96; 100 / 100 |
| Warm lighting, strong | 64; 66.7 / 80.0 | 64; 66.7 / 80.0 | 94; 97.9 / 98.9 |
| Cool lighting, moderate | 96; 100 / 100 | 75; 78.1 / 87.7 | 96; 100 / 100 |
| Cool lighting, strong | 49; 51.0 / 67.1 (FP 1) | 18; 18.8 / 31.0 (FP 2) | 94; 97.9 / 98.9 |
| Under-exposure −1 EV | 96; 100 / 100 | 96; 100 / 100 | 96; 100 / 100 |
| Under-exposure −2 EV | 96; 100 / 100 | 96; 100 / 100 | 96; 100 / 100 |
| Over-exposure +0.65 EV | 96; 100 / 100 | 96; 100 / 100 | 96; 100 / 100 |
| Channel mixing 0.10 | 96; 100 / 100 | 96; 100 / 100 | 96; 100 / 100 |

| Condition | Initial colour update (rule-based) | Colour adaptation (rule-based) | VLM guidance |
|---|---|---|---|
| Channel mixing 0.25 | 96; 100 / 100 | 65; 67.7 / 80.7 | 96; 100 / 100 |
| Sensor noise, moderate | 96; 100 / 100 | 96; 100 / 100 | 96; 100 / 100 |
| Sensor noise, strong | 96; 100 / 100 | 96; 100 / 100 | 96; 100 / 100 |
| Defocus blur r = 1 pixel | 96; 100 / 100 | 96; 100 / 100 | 96; 100 / 100 |
| Defocus blur r = 2 pixels | 96; 100 / 100 | 96; 100 / 100 | 96; 100 / 100 |
| JPEG q = 60 | 96; 100 / 100 | 96; 100 / 100 | 96; 100 / 100 |
| JPEG q = 30 | 96; 100 / 100 | 96; 100 / 100 | 96; 100 / 100 |
| Combined: warm lighting + under-exposure + noise | 96; 100 / 100 | 96; 100 / 100 | 96; 100 / 100 |
| Combined: cool lighting + printing deviation + channel mixing | 15; 15.6 / 26.5 (FP 2) | 15; 15.6 / 26.5 (FP 2) | 64; 66.7 / 80.0 |
| Combined: under-exposure + channel mixing + noise + JPEG | 95; 99.0 / 99.0 (FP 1) | 93; 96.9 / 96.9 (FP 3) | 96; 100 / 100 |
| Total (2,016) | 1855; 92.0 / 95.7 (FP 4) | 1770; 87.8 / 93.3 (FP 7) | 1980; 98.2 / 99.1 (FP 0) |

### 3.4 Processing time

After a profile was obtained, the per-frame tracker processing time averaged 395 ms for Initial colour update, 373 ms for Colour adaptation and 286 ms for VLM guidance; VLM guidance was the shortest, but the three methods were of the same order of magnitude. The two rule-based methods appear to have been slightly slower because the tracker searched a wider range in the many frames in which detection failed, but the processing time was not decomposed by stage. These are single-thread comparative values; the deployed program runs faster with multiple threads. With the model already loaded on the GPU, VLM guidance takes about 15 s to locate the markers in one frame and about 25 s including the colour acquisition from that frame and its native detection support. Because the profile is reused after the two start-up costs, model loading and the first colour acquisition, the tracking speed is no different from that of the rule-based methods. Both training and evaluation were done on the same single GPU (NVIDIA GeForce RTX 4070 Laptop GPU, 8 GB).

*Estimated VLM cost by GPU class.* The VLM cost depends on the GPU class. When a 4-bit quantised 4B model is run with batch size 1, token generation is bound mainly by GPU memory bandwidth, and the prefill of the image tokens by compute performance. The memory bandwidth of the RTX 4070 Laptop GPU used for the measurements is 256 GB/s, that of the desktop RTX 4090 is about 1 TB/s, and those of the data-centre A100 and H100 are about 2 TB/s and 3.4 TB/s. The ratios of compute performance are similar or larger. The same code is therefore estimated to run 3–5 times faster on a desktop GPU and 6–10 times faster on a data-centre GPU, which would reduce the time of about 25 s per frame to about 5–8 s and 2.5–4 s respectively. Colour acquisition within 10 s is therefore readily achievable on a desktop GPU. On a GPU with 24 GB or more of memory the model can run in 16 bits without quantisation, which also removes the de-quantisation cost. Model loading is governed by the storage and PCIe speed and is of the order of several seconds to 20 s regardless of GPU class. These estimates are based on published specifications and are not measurements.

## 4 Discussion

The three methods complement rather than replace one another. The two rule-based methods need no training and no GPU, start immediately and are easy for a person to follow and check; they are sufficient when the camera and the lighting are similar to those at design time. Colour adaptation suits long sessions in which the illumination changes slowly, although an update in the wrong direction can become frozen, so validation and reversion must accompany it.

The greatest weakness of the rule-based methods is that both lose their starting point when the recorded colours deviate greatly from the marker definition. The reason is that they are bound to the circle in which the marker must be found with the design colours before the colours can be corrected. VLM guidance obtains the first candidate and its ID from the learned appearance, so it stands outside this circle, and this difference contributes very effectively when the colour change is large. The result on the surgical-robot video shows the advantage of VLM guidance most clearly. In this video the purple stripe of CS appeared with the same hue (about 220°) as the blue stripe of RS, and the saturation of the orange stripe fell to one third of the design value (Figure 1(b)). The two rule-based methods never found the CS marker, whereas VLM guidance found all 234 markers. This was possible because VLM guidance fed the correct pixels into the same profile rules even for a marker on which the design-colour detection never succeeded. It is reasonable to conclude that the gain on videos with large colour changes comes from a VLM candidate search that does not rely on the design colours.

What this paper shows is how reliably a method fixed on non-target development data obtains the first colour sample under the same processing after the colours are obtained, and that under conditions in which design-colour initialisation is difficult, VLM guidance gave a higher recognition rate than the compared rule-based methods.

The cost of VLM guidance is a GPU and the delay while the colours are acquired, about 25 s per frame on the present laptop GPU; a desktop GPU is expected to be at least three times faster. This cost is paid once at the start of tracking, and the profile is reused afterwards, so tracking does not slow down. As described above, the hues of two different colour pairs overlapped in the surgical-robot video. In a design in which the colour pair also serves as the ID, such an overlap can lead to ID confusion, so in a site with fixed lighting, the combination of colour pairs should also be chosen to suit that lighting.

The simulations also show the strengths and weaknesses of the marker structure and of the colour information in a slender mounting area. The large features and the long centre line of the stripe structure apply to both black-and-white and colour markers.

A representative case is lip protection in TORT. In TORT, three trocars are inserted through incisions inside the lower lip, and the instruments are moved about a remote centre of motion (RCM) [37,38]. The point where the trocar touches the lip is 1–3.5 cm from the RCM, so the lip is rubbed every time an instrument pivots. On a mannequin, the displacement of the lip point measured with an optical tracker was 5–21 mm at P95 and was proportional to the distance between the lip point and the RCM. Mental nerve injury, laceration of the corner of the mouth and skin burns are complications characteristic of TORT [1,2,39]. Monitoring the trocar pose with a camera can give a warning before the force on the lip becomes large. These trocars are 10 mm in diameter, a little over 3 cm apart and partly hidden by the lip and the robot arm, so a small narrow marker with colour calibration adapted to the operating-room lighting is very useful. The same effect can be expected for laparoscopic trocars, endoscopes and electrocautery tools, of which only the handle end is exposed.

This study has some limitations. The recognition rate concerns the position and ID of the markers in the two-dimensional image; the accuracy of the three-dimensional pose was not measured, but it can be judged from the laboratory measurements of the previous study [15] and from the simulation in Table 4. The evaluation annotations and detection judgements of the laboratory and surgical-robot videos were prepared with the assistance of a separate, recent large language model (GPT-6 Astra, OpenAI), and the authors then verified some of the samples against the videos. The surgical-robot result comes from a single recording on a mannequin. With a dedicated camera and a higher-performance GPU, a larger model becomes possible and the image quality and processing time can improve. Even with the present 8 GB GPU, however, VLM guidance surpassed the two rule-based methods on every data set, and the remaining failures appear to lie in the processing after the model rather than in the model itself.

## 5 Conclusion

The colour change that is the weakness of colour-stripe markers can be compensated by sampling pixels from the real-time input and updating the profile. The rule-based methods are sufficient when the degradation is mild. In the surgical-robot video, however, they never found the marker whose colours had shifted severely, whereas VLM guidance found all markers. A VLM is not necessary under all conditions, but when design-colour initialisation is difficult it adds an effective option for an automatic pipeline in which a person does not have to re-adjust the colours. The proposed adaptive colour calibration opens the way to tracking slender surgical instruments with small colour markers.

## Declarations

**Funding.** This work was supported by the National Research Foundation of Korea (NRF) grant funded by the Korea government (MSIT) under Grant RS2024-00355535.

**Disclosure statement.** The authors report there are no competing interests to declare.

**Declaration of generative AI use.** The authors determined the scientific content, the algorithms of the original methods and the manuscript organisation. Claude Fable 5.2 (Anthropic) assisted with text editing and figure design. ChatGPT/Codex (GPT-6 Astra, OpenAI) assisted with algorithm implementation, simulation and evaluation code. The authors reviewed all AI-assisted materials, verified the computational outputs and revised the manuscript and figures. Experimental data and photographs came from the authors' experiments, and plotted values were derived from measurements or numerical simulations. Generative AI was not used to fabricate measurements or synthesise experimental photographs. The authors have checked the terms of use of these tools and their suitability for publication, have confirmed the originality and accuracy of the content, and take full responsibility for the final manuscript, including the accuracy of the references.

**Data availability statement.** The data supporting the findings of this study are available from the corresponding author upon reasonable request. The data are not publicly available to protect intellectual property associated with potential future patent applications.

**Ethics statement.** This study used a mannequin and bench-top set-ups only and did not involve human participants, patients or animals; ethical approval was therefore not required.